\documentclass{article}

\usepackage{microtype}
\usepackage{graphicx}
\usepackage{subfigure}
\usepackage{booktabs} 

\usepackage{breakurl}
\usepackage[breaklinks=true]{hyperref}

\usepackage[accepted]{icml2024}

\usepackage{amsmath}
\usepackage{amssymb}
\usepackage{mathtools}
\usepackage{amsthm}

\usepackage[capitalize,noabbrev]{cleveref}

\theoremstyle{plain}

\theoremstyle{definition}

\theoremstyle{remark}

\usepackage[textsize=tiny]{todonotes}

\icmltitlerunning{The Tug-of-War Between Deepfake Generation and Detection}

\begin{document}

\twocolumn[
\icmltitle{The Tug-of-War Between Deepfake Generation and Detection}




\begin{icmlauthorlist}
\icmlauthor{Hannah Lee}{uw,truemedia}
\icmlauthor{Changyeon Lee}{yonsei,purdue,miraflow}
\icmlauthor{Kevin Farhat}{uw,truemedia}
\icmlauthor{Lin Qiu}{uw,truemedia}
\icmlauthor{Steve Geluso}{truemedia}
\icmlauthor{Aerin Kim}{miraflow,truemedia}
\icmlauthor{Oren Etzioni}{truemedia}
\end{icmlauthorlist}

\icmlaffiliation{uw}{Paul G. Allen School of Computer Science \& Engineering, University of Washington, Seattle WA, USA}
\icmlaffiliation{purdue}{Computer and Information Technology, Purdue University, West Lafayette IN, USA}
\icmlaffiliation{truemedia}{TrueMedia.org, Seattle WA, USA}
\icmlaffiliation{miraflow}{Miraflow, Kirkland WA, USA}
\icmlaffiliation{yonsei}{Department of Computer Science and Engineering, Yonsei University, Seoul, Republic of Korea}

\icmlcorrespondingauthor{Hannah Lee}{hannahyk@cs.washington.edu}

\icmlkeywords{machine learning, deepfakes, deepfake detection, deep learning, generative models, misinformation, survey, generative adversarial networks, diffusion, datasets}

\vskip 0.3in
]



\printAffiliationsAndNotice{}  

\newcommand{\edit}[1]{\textcolor{red}{#1}}
\newcommand{\assigned}[1]{\textcolor{blue}{#1}}

\begin{abstract}
Multimodal generative models are rapidly evolving, leading to a surge in the generation of realistic video and audio that offers exciting possibilities but also serious risks. Deepfake videos, which can convincingly impersonate individuals, have particularly garnered attention due to their potential misuse in spreading misinformation and creating fraudulent content. This survey paper examines the dual landscape of deepfake video generation and detection, emphasizing the need for effective countermeasures against potential abuses. We provide a comprehensive overview of current deepfake generation techniques, including face swapping, reenactment, and audio-driven animation, which leverage cutting-edge technologies like generative adversarial networks and diffusion models to produce highly realistic fake videos. Additionally, we analyze various detection approaches designed to differentiate authentic from altered videos, from detecting visual artifacts to deploying advanced algorithms that pinpoint inconsistencies across video and audio signals.

The effectiveness of these detection methods heavily relies on the diversity and quality of datasets used for training and evaluation. We discuss the evolution of deepfake datasets, highlighting the importance of robust, diverse, and frequently updated collections to enhance the detection accuracy and generalizability. As deepfakes become increasingly indistinguishable from authentic content, developing advanced detection techniques that can keep pace with generation technologies is crucial. We advocate for a proactive approach in the  ``tug-of-war'' between deepfake creators and detectors, emphasizing the need for continuous research collaboration, standardization of evaluation metrics, and the creation of comprehensive benchmarks.

\end{abstract}

\section{Introduction}
\label{introduction}
Recent advances in multimodal generative models have made manipulated media increasingly more realistic and accessible. Although synthetically generated audio, images, and videos can have creative and beneficial applications such as improved dubbing or translation of films
\cite{yang2020large, neuraldubber}, deepfake videos that impersonate humans highlight the potential harms of media manipulation and synthetic generation. For example, deepfakes that blend celebrities' faces onto bodies in pornographic videos \cite{Nguyen_2022, taylorswift-fakes} and alter politicians' messages \cite{suwajanakorn2017synthesizing} can spread misinformation, threaten individuals, and damage reputations \cite{Zhou_2020}, disrupting election campaigns and financial markets. Recently, deepfakes have also become integral to fraudulent schemes, resulting in scams of up to \$25 million \cite{chen2024finance}. Given the rise of social media and online media consumption, it is unsurprising that deepfake videos are increasingly interfering with people's lives.

The first modern deepfake videos surfaced in 2017 when users on Reddit posted computer-generated pornographic videos of actresses \cite{Nguyen_2022}. Since then, many deepfake generation tools have become available for public use. FaceSwap \cite{FaceSwap}, FaceSwapGAN \cite{faceswap-gan}, StyleGAN \cite{karras2019style}, FSGAN \cite{FSGAN} and other applications allow anyone with basic programming skills to generate their own deepfakes, and as text-to-video models such as Imagen Video \cite{ho2022imagen}, CogVideo \cite{hong2022cogvideo}, and Sora \cite{videoworldsimulators2024} improve and become widespread, the barrier to entry will continue to be lowered. 

As deepfake video generation has become increasingly democratized and capable of producing realistic results, emphasis on countermeasures has also grown. Generally, addressing misinformation and the harms related to deepfakes follows two patterns: detection and prevention. While prevention through techniques like watermarking \cite{lv2021smart} and blockchain frameworks \cite{rashid2021blockchain} as well as through technology policy \cite{reisach2021responsibility} is critical for mitigating deepfake harms, it is out of scope for this paper and we instead focus on deepfake detection, the currently more common strategy. 

Deepfake video detection is comprised of fake image detection, fake audio detection, and techniques specific to video sequences. These techniques are generally leveraged for the binary classification task where classifiers learn to distinguish authentic and manipulated videos. Initial detection tools focused primarily on visual artifacts such as blended face edges \cite{li2020face} and image forgery techniques that have preceded deepfakes \cite{Chaitra2022DigitalIF}. More recently, deep learning techniques have capitalized on the large amounts of real and fake data available online. However, since training detection algorithms depends on fake data created by generation tools, deepfake detectors lag behind generators. Consequently, the development of detection algorithms provides direct feedback to generation algorithms on what makes deepfakes detectable and can encourage adversarial generation to bypass detection. We refer to this relationship as the ongoing tug-of-war between deepfake generation and detection.

There are existing survey papers that explore deepfake generation and detection \cite{rana2022detection, Yu2021detection, yi2023audio, Mirsky_2021, Nguyen_2022, swathi2021deepfake}. However, these often only focus on detection \cite{rana2022detection, Yu2021detection, yi2023audio}; do not include modern methods that have gained popularity \cite{Mirsky_2021, swathi2021deepfake}; or do not consider in depth the impact of the datasets used to train generation and detection models. In our approach, we focus on video generation and detection while discussing the current landscape of curating relevant training and evaluation datasets. Our main contributions include an updated survey of deepfake generation and detection techniques; identification of the successes, challenges, and limitations of current deepfake detection practices; and suggestions for future deepfake detection research, focusing on the importance of quality datasets in this tug-of-war.

\section{Deepfake Video Generation}
\label{generation}
Generating deepfake video media consists of generating both visual and audio content. Here, we briefly describe common generation processes for both modalities.

\subsection{Face Swapping} 
 Face swapping is one method used to create deepfake images and videos by replacing one person's face with another's face. This technique is now widely available through well-developed packages like Face Fusion~\cite{facefusion_software} and Faceswap~\cite{FaceSwap}. Early works could only handle faces in the same pose, but later developments have incorporated 3D-based methods to construct faces with variations~\cite{Dale2011VideoFR, Lin2012FaceSU}. With the development of neural networks, face swapping can now involve an encoder-decoder network~\cite{DeepFaceLab}. The encoder extracts latent features from the faces in the source image while the decoder reconstructs the target image using these features. More recently, GAN-based methods have also been applied; FSGAN~\cite{FSGAN} and FSGAN2~\cite{nirkin2022fsganv2} provide a two-stage pipeline that supports both face swapping and reenactment (further discussed in Section \ref{generation:reenactment}) simultaneously. These methods train a generator to learn the latent representation of the source image and then inpaint the segmented area in the target image. These GAN-based methods are able to generate more realistic images with higher resolution and fidelity. However, training GANs can be unstable, restricting their actual application to lower resolution images~\cite{xu2022highresolution, liu2023finegrained}.

\subsection{Reenactment} \label{generation:reenactment} 
Reenactment focuses on making one person's face in a video mimic the facial expressions and head movements of another person. Unlike face swapping, the identity of the face remains the same while its expressions and movements are altered~\cite{Nguyen_2022}. Reenactment can drive the expression, gaze, mouth, pose, identity, and body of the target subject based on the source subject~\cite{Mirsky_2021}. Face2Face~\cite{thies2016face2face} proposed the first method that achieves real-time RGB-only reenactment without the need for a teeth proxy~\cite{automaticfacereenactment, thies2015realtime} or direct source-to-target copying~\cite{faceMutilinearModel}. FSGAN can be applied to unseen pairs of faces and adjusts significant pose and expression variations that can be applied to a single image or a video sequence~\cite{FSGAN}. Later, FSGAN2 extended the model by introducing preprocessing and additional post-processing steps for reducing flickering and saturation artifacts~\cite{nirkin2022fsganv2}. Recently, facial reenactment efforts have not only focused on updating the expressions on the target image itself but have also involved a joint effort in creating both audio and visual components for talking face generation (Section \ref{sec:audio-driven-generation}).

\subsection{Diffusion-based Deepfakes} 
Diffusion-based deepfake generation represents a significant advancement over traditional methods like GANs and autoencoders, offering more realistic and efficient capabilities for generating deepfakes. Recent developments in diffusion models (DMs), such as Denoising Diffusion Probabilistic Models (DDPM)~\cite{Ho2020DenoisingDP}, have further refined this approach. DDPM utilizes a series of learned noise adjustments to reverse the addition of noise, effectively reconstructing the data distribution of original images. Subsequently, the Latent Diffusion Model (LDM)~\cite{Rombach2021HighResolutionIS} enhances this process by generating in the latent space, speeding up the diffusion process and handling complex data distributions. 

\noindent\textbf{Image Generation.}
Diffusion-generated images are categorized into two types: text-to-image, where users provide text prompts to specify image content to models such as DALL-E~\cite{Ramesh2021ZeroShotTG}, DreamBooth~\cite{Ruiz2022DreamBoothFT} and Imagen~\cite{Saharia2022PhotorealisticTD}; and text guided image editing or generation, which transforms existing images based on text prompts using Imagic~\cite{Kawar2022ImagicTR}, InstructPix2Pix~\cite{Brooks2022InstructPix2PixLT}, or other tools. Chen et al.~\citeyearpar{Chen2023TextimageGD} proposes a method to specifically generate high quality deepfakes by leveraging diffusion, using celebrity images as a guide for the image's latent initialization.

\noindent\textbf{Video Generation.} 
DMs are also capable of generating videos. Typically, this is done by inflating 2D convolutional layers into pseudo-3D convolutional layers in the DM and incorporating temporal self-attention layers in each transformer block to manage the temporal consistency across frames. As one example, Diffusion Heads~\cite{Stypulkowski2023DiffusedHD} leverages an autoregressive DM that takes one identity image and an audio sequence to generate a talking head, which can also be applied to the deepfake generation with a custom identity image. 

\subsection{Audio-Driven Facial Animation}\label{sec:audio-driven-generation} 

The field of audio-driven facial animation has been a fascinating area of exploration for the computer vision and graphics community. Many studies have been carried out on both digital 3D human faces~\cite{karras2017audio,richard2021meshtalk,zhou2018visemenet,fan2022faceformer} and realistic human head generation~\cite{fan2015photo,suwajanakorn2017synthesizing,jamaludin2019you}. 

\noindent\textbf{Talking Head Generation.}
Early research on lip-syncing mostly focused on creating the entire head of a speaking person~\cite{suwajanakorn2017synthesizing,jamaludin2019you,kaisiyuan2020mead,zhang2023dinet}. Several methods used structural details like 2D~\cite{chen2019hierarchical} landmarks, 3D landmarks~\cite{zhou2020makelttalk} and 3D meshes~\cite{chen2020comprises}. However, the resulting output often exhibits inconsistencies between connecting parts and suffers from a lack of precision. Many methods that are specific to individuals ~\cite{suwajanakorn2017synthesizing,lu2021live,ji2021audio} depend on the subject and may struggle with generalization. Furthermore, the recent application of NeRF for person-specific modeling, as cited in several works~\cite{guo2021adnerf,liu2022semantic,shen2022dfrf}, often exhibits sub-optimal performance, including low visual quality, restricted expressiveness, and inconsistency across frames, when the available training data is limited.

\noindent\textbf{Lip-Syncing on Faces.}
Other research has focused on lip-syncing the mouth in videos while leaving other elements unchanged~\cite{prajwal2020lip,park2022synctalkface,thies2020neural}. Specifically, Wav2Lip~\cite{prajwal2020lip} is capable of generating lip-sync results that are not person-specific. However, the generative model used is built on low-resolution images, resulting in blurry outputs. It also falls short in capturing the unique identity of a target including the shape of teeth and lips when provided with a target template video.

\subsection{Style-based Generator for Faces}
StyleGAN models~\cite{karras2019style,Karras2019stylegan2,Karras2021} have shown great success on image generation tasks, particularly on facial image generation and editing~\cite{abdal2019image2stylegan,abdal2020image2stylegan++,tov2021designing,richardson2021encoding}. 
They have also been leveraged for face restorations~\cite{wang2021gfpgan,yang2021gan} and face swapping~\cite{xu2022styleswap, xu2022region}, which require the preservation of the original facial emotion and expressions. Concurrently, style-based generators have been effectively utilized in the creation of facial animations~\cite{burkov2020neural,liang2022expressive}. However, the generators rely on the style vectors in the $W$ or $W^+$ latent spaces for controlling both the appearances and motion dynamics. A significant limitation of this approach is the $W^+$ space's inability to maintain the spatial consistency of backgrounds. This often results in non-realistic outcomes or noticeable artifacts.


\subsection{Audio Generation} 
Deepfake audio generation involves creating realistic synthetic audio, often mimicking a specific person's voice. We describe common techniques used to produce highly convincing audio deepfakes.

\textbf{Text-to-Speech (TTS).} Text-to-speech systems convert written text into spoken words. Modern TTS systems use deep learning models to produce natural-sounding speech. Key approaches include:
\begin{itemize}
    \item \textbf{Concatenative TTS:}
A traditional method involving concatenating pre-recorded speech segments to generate fluent speech. While it can produce high-quality and natural-sounding results by leveraging real audio, it often struggles with capturing the dynamic variations in human speech~\cite{hande2014volume}. This can lead to noticeable discontinuities and a lack of flexibility when synthesizing speech with varying intonation, emphasis, or speaking styles~\cite{cohn20_interspeech}.
    \item \textbf{Parametric TTS:} Generating speech by predicting a set of parameters that describe aspects of the speech signal such as pitch, duration, and spectral features by using statistical models like Hidden Markov Models~\cite{Tokuda2000SpeechPG}. This approach offers flexibility in creating various voices and speaking styles, but the synthesized speech may sound less natural due to the oversimplification of speech variations.

    \item \textbf{Neural TTS:} Using deep learning models to generate more natural and expressive speech~\cite{tan2021survey}. Examples include Tacotron 2~\cite{Shen2017NaturalTS} and WaveNet~\cite{Oord2016WaveNetAG}, where Tacotron 2 converts text to a mel-spectrogram then uses a modified WaveNet as a vocoder to generate audio waveforms from the spectrogram.
\end{itemize}

\textbf{Voice Conversion (VC).} VC aims to modify a source speaker's voice to sound like a target speaker without changing the linguistic content. This process typically involves extracting features from the source speech, such as pitch, spectral envelope, and timbre, then transforming these features to match the target speaker's characteristics. Statistical models like Gaussian Mixture Models (GMMs)~\cite{Kain1998SpectralVC} and deep learning approaches such as GANs and variational autoencoders (VAEs) are employed to learn the mapping between the source and target features. Recent advancements have significantly improved the naturalness and intelligibility of converted speech~\cite{Sisman2020-qq}.

\textbf{Emotion Fake}. Emotion fake, or emotional VC, involves altering the emotional tone of a speaker's voice. This technique is essential for creating more expressive and engaging synthetic speech. It typically involves extracting prosodic features such as pitch, duration, and intensity before modifying these features to reflect the desired emotion. Deep learning methods, particularly Convolutional Neural Networks (CNNs), GANs and Recurrent Neural Networks (RNNs), have shown promising results in this area~\cite{VanESR, Liu2020EmotionalVC}. These models can learn complex mappings between neutral and emotional speech, resulting in more natural and convincing emotional expressions. 

\textbf{Scene Fake.} Scene fake, also known as environmental sound synthesis, involves generating background sounds or environmental effects to accompany synthetic speech to enhance the overall experience. This technique aims to create a more realistic auditory scene by simulating various ambient sounds such as street noise, office chatter, or natural environments like piano or birdsong~\cite{donahue2019adversarial, huzaifah2020deep}. GANs and autoencoder-based architectures have been applied to synthesize scene fake audio. 

\textbf{Partially Fake.} Partially altered audio involves modifying specific words or segments within a spoken recording by replacing them with either authentic or artificially generated audio snippets. The speaker's voice remains consistent throughout the original speech and the altered segments, making the modified audio sound authentic despite the introduced changes. This technique allows for counterfeit versions of the initial speech to be created while maintaining the speaker's identity and vocal characteristics~\cite{Wu2022PartiallyFA}.

\section{Deepfake Video Detection}
\label{detection}
Deepfake video generation techniques necessitate a broad arsenal of detection tools. Fake video detection is generally broken down into three main categories; fake image detection where individual frames are analyzed; fake audio detection to analyze the audio of a video; and fake video detection, which may utilize both images and audio as well as temporal data.

\subsection{Fake Image Detection}

Detecting fake images precedes deepfakes and deep learning. We discuss methods ranging from artifact detection to modern methods that have developed with advancements in deep learning.  

\textbf{General Visual Artifact Detection.}
Deepfake images may introduce subtle artifacts that are not present in real images. These can include artifacts introduced by face blending or face warping when swapping faces, as well as inconsistencies in the overall image. For example, Li et al.~\citeyearpar{li2020face} propose the ``face X-ray'' image representation to detect anomalies in the blending boundaries of faces in blended images. Other work has focused on the differing image textures between generated and real content~\cite{liu2020global}; inconsistencies in head positions~\cite{yang2019exposing}; the resolution variability that arises from warping faces~\cite{li2018exposing}; and missing reflections or details in the teeth and eyes~\cite{matern2019exploiting}. Many other visual artifact detection methods have been proposed, but as deepfake generation has improved and fewer artifacts remain, these techniques have become overshadowed by methods focused on detecting more subtle identifiers.

\textbf{Detecting GAN-generated Images.}
A subset of image deepfake detection methods have focused on detecting artifacts unique to popular GAN models. Some detection methods still rely on visible differences such as irregular pupil shapes~\cite{guo2022eyes}, while many others exploit lower level abnormalities. For example, the upsampling operations involved in GAN generation introduce model specific artifacts into the images' spatial and frequency domains~\cite{zhang2019detecting, marra2019gans, Yu2021detection}. Wang et al.~\citeyearpar{wang2019cnngenerated} trained a ResNet-50 model \cite{he2016deep} that detects these artifacts and found that GAN-generated images are easily detected. Similarly, FakeSpotter~\cite{wang2019fakespotter} can effectively detect AI-synthesized fake faces generated by GANs. Building upon these detectors, PatchForensics~\cite{chai2020makes} introduced a detector that analyzes smaller patches of images to determine if there are AI-generated or manipulated areas.

\textbf{Diffusion Detection.}
Generalization of GAN-based detection methods to newer, diffusion based image generation techniques is difficult~\cite{corvi2023detection, ojha2023towards}. The artifacts introduced by GANs in images are no longer present with DMs. Recently, Wang et al.~\citeyearpar{wang2023dire} proposed DIRE, a new image representation that uses reconstructions of images using DMs as a method for detecting DM generated images. They hypothesize that DM generated images consist of features that are better reconstructed by other pretrained DMs, compared to reconstructions of real images. Furthermore, Lim et al. \citeyearpar{lim2024distildiresmallfastcheap} introduced DistilDIRE, a diffusion-generated image detection framework that significantly reduces the computational demands of the original DIRE method.  Ojha et al.~\citeyearpar{ojha2023towards} instead utilize the learned feature space of a pretrained vision-language model to determine if an image was AI-generated. And Lorenz et al.~\citeyearpar{lorenz2023detecting} rely on multi Local Intrinsic Dimensionality to detect diffusion. These novel methods highlight how detection algorithms continue to adapt to the newer generative models.

\subsection{Fake Audio Detection}

Audio deepfake detection can utilize traditional classifiers like Support Vector Machines (SVM) after feature extraction using methods such as STFT spectrograms or Mel-frequency Cepstral coefficients (MFCC). However, deep learning models outperform traditional classifiers by learning complex patterns, resulting in better accuracy; these approaches are now widely preferred~\cite{10258355}.

Typically, deep learning-based audio deepfake detection involves extracting image features like STFT spectrogram feature images, followed by employing a CNN-based architecture to extract embeddings. A binary classification layer is then added to classify as real or fake. However, other techniques that are often employed in traditional audio classification can also be applied. RNN-based models can capture temporal patterns of audio signals due to their sequential nature~\cite{10258355}, enabling classification into different categories~\cite{9287802, Gimeno2020}.

More recently, transformer-based models have been introduced. These models, when compared to CNN-based ones, can handle input-length variance due to their multi-head self-attention mechanisms~\cite{10258355}. This allows them to effectively capture useful global-context information, regardless of the audio length~\cite{gong2021ast, ghosh2023mast, liu2023cat}.

\subsection{Fake Video Detection}

Deepfake detection techniques for videos transcend detection techniques for single images by comparing images temporally frame after frame. Deepfake video detection techniques may also combine information from images and audio to make inferences based on coherence, synchronization, and physiology.

\textbf{Frame by Frame Analysis.}
Evidence of deepfake video generation may be apparent analyzing videos frame by frame. Determining movement via optical flow may distinguish real videos from fake videos~\cite{amerini2019opticalFlow}. Guera and Delp~\citeyearpar{guera2018recurrent} use a CNN and RNN to extract features in frames and detect inconsistencies from the overall frame sequence, acknowledging deepfake manipulations often only appear briefly in overall videos. Taking one frame, and predicting the next frame, and measuring the error between the prediction and the actual next frame provides a basis for prediction error energy analysis~\cite{amerini2020predictionError}. Frame analysis techniques can be improved by pre-processing video data to focus on area where faces occur~\cite{Sabir2019recurrentFaceDetection}.

Recent frame by frame methods include GenConViT~\cite{wodajo2023genconvit}, which generalizes deepfake video detection by extracting latent spaces of video frames using two networks with an independently trained autoencoder and VAE. The VAE reconstructs images and compares the reconstruction to the sample image. The autoencoder and VAE each feed into a ConvNeXt layer, and a Swin Transformer forming a hybrid model ConvNeXt-Swin which learns the relationships among the latent features. Sun et al.~\citeyearpar{sun2023faceforgery} also analyzes frames, tracking the horizontal and vertical displacement trajectories of virtual anchor points on faces over time. They found real videos produce smoother trajectories than anchor points on fake faces and developed a fake trajectory detection network to classify real and fake videos. 

\textbf{Physiological Features.} Generative techniques often lack artifacts of biological processes. Deepfake video detection techniques have been able to identify deepfakes by analyzing how people in the video blink \cite{li2018InIctuOculi}, whether the shape of their mouth is in synchronization with the sounds their mouth is making \cite{agarwal2020PhonemeVisemeMismatches}, and measuring heart rate via photoplethysmography (PPG) signals from hemoglobin content in the blood changing how skin reflects light \cite{ciftci2020heartbeat}. 

\textbf{Audio-visual Analysis.} Some work has observed inconsistencies between images and audio. Chugh and Subramanian~\citeyearpar{chugh2021audiovideodissonance} observe that deepfake videos often have dissonance between the audio and video and create a metric called the Modality Dissonance Score. Computing audio-visual dissimilarity over 1-second video clips, they then aggregate these scores to perform classification.

\subsection{Adversarial Attacks and Evading Detectors}

Despite numerous detection methods, studies have found that evasion efforts can be effective at rendering some detectors useless. Carlini and Farid~\citeyearpar{carlini2020evading} found multiple small perturbation attacks that can be applied to deepfake images, resulting in the performance of the classifier by Wang et al.~\citeyearpar{wang2019cnngenerated} to be reduced to worse than chance while minimizing distortions visible to humans. Hou et al.~\citeyearpar{hou2023evading} and Neekhara et al.~\citeyearpar{neekhara2021adversarial} build on this work to extend the attacks and apply to other detectors. While these works do not include evasions of more modern detection methods, they speak to the tug-of-war nature between generation and detection.

\section{Detection Challenges, Competitions, and Datasets}
\label{datasets}
Each of the methods described in 
Sections \ref{generation} and \ref{detection} have been trained on datasets of curated real and fake images. Here, we discuss the current landscape of available datasets and the progression of modern deepfake video detection algorithms. Table \ref{tab:datasets} summarizes some commonly used, publicly available datasets.

\begin{table*}[t]
\centering
    \caption{Summary of common datasets used for training and evaluating deepfake detection models.}
\vskip 0.15in
\begin{tabular}{ccccccc}
\toprule
Dataset                             & Modality & Identities & Real Samples     & Generated Samples  & Generation Methods                             & Year \\
\midrule
CNNDetect                           & Image    & /          & 72,400                & 72,400               & Multiple CNNs                                  & 2020 \\
CIFAKE                              & Image    & /          & 60,000                & 60,000               & Diffusion                                      & 2024 \\
\hline
FoR                                 & Audio    & /          & \textgreater{}111,000 & \textgreater{}87,000 & TTS                                            & 2019 \\
ASVspoof (LA)                    & Audio    & 107        & 12,483                & 108,978              & TTS, VC                                        & 2019 \\
H-Voice                             & Audio    & /          &  3,268                & 3,404                & Multiple                                       & 2020 \\
WaveFake                            & Audio    & /          & /                    & 117,985             & Multiple                                       & 2021 \\
In-the-Wild                         & Audio    & 58         & 20.7 hours           & 17.2 hours          & Multiple                                       & 2022 \\
EmoFake                             & Audio    & 10         &  17,500               & 36,400               & Multiple EVC Models                            & 2022 \\
SceneFake                           & Audio    & 107        &  19,838               & 64,642               &  Multiple                                      & 2022 \\
DEEP-VOICE                          & Audio    & 8          & 62 min 22 sec        & 62 min 22 sec       & RVC model                                      & 2023 \\
ADD                                 & Audio    & /          & 243,194               & 273,874              & Multiple                                       & 2023 \\
\hline
DeepfakeTIMIT                       & Video    & 32         & 320                  & 640                 & GAN (face swap)                                & 2018 \\
FaceForensics++                     & Video    & /          & 1,000                 & 4,000                & Multiple                                       & 2019 \\
Celeb-DF                            & Video    & 59         & 590                  & 5,639                & GAN (face swap)                                & 2019 \\
WildDeepfake                        & Video    & 707        &  3,805                &  3,509               &  Multiple                                      & 2020 \\
DFDC                                & Video    & 960        & 23,654                & 104,500              & Multiple                                       & 2020 \\
DeeperForensics-1.0                 & Video    & 100        &  50,000               & 10,000               & DF-VAE (face swap)                             & 2020 \\
AV-Deepfake1M                       & Video    & 2,068       & 286,721               & 860,039              &  Multiple                                      & 2023  \\
\bottomrule
\end{tabular}
    \label{tab:datasets}
\end{table*}

\subsection{Deepfake Image Datasets}

Many datasets of real faces already exist for facial recognition tasks. In 2007, the Labeled Faces in the Wild (LFW) database curated over 13,000 images of faces from the internet~\cite{LFWTech}. Though now retracted, the MS-Celeb-1M dataset consisted of one million real images of 100,000 different identities~\cite{guo2016ms}. And IMDB-WIKI~\cite{Rothe-IJCV-2018} and IMDB-Clean~\cite{lin2021fpage} include 524,230 and 287,683 images respectively of faces, designed to train age-estimation algorithms. These datasets also aid the training of face image generation models.

For deepfake image generation techniques that rely on face swapping or reenactment, the deepfake creator must have sufficient training data focused on both subjects~\cite{FaceSwap, facefusion_software}. This has limited deepfakes to be of celebrities with large amounts of publicly available content, but as the quantity of public data increases and generation techniques improve, the accessibility to creating deepfakes continues to rise. The CelebA-HQ dataset~\cite{karras2017progressive} is one early dataset of 30,000 images of the faces of celebrities, based on the larger CelebA dataset~\cite{liu2015faceattributes}, that has been used to train GANs to generate higher resolution images of human faces. Karras et al.~\citeyearpar{karras2019style} went on to create Flickr-Faces-HQ (FFHQ), a larger and more diverse dataset of 70,000 images no longer limited to celebrities, inclusive of broader ranges of ages, ethnicities, and image backgrounds.

In contrast to deepfake generation models, detection models are typically trained on large datasets of both real and fake images. Often, these are custom datasets depending on the detection focus and currently available image generators. Wang et al.~\citeyearpar{wang2019cnngenerated} trained an image classifier on the LSUN~\cite{yu2015lsun} dataset and ProGAN~\cite{karras2019style} generated images. To test their image classifier, Wang et al. generated over 72,000 images using 11 different CNN-based models including StyleGAN~\cite{karras2019style}, BigGAN~\cite{brock2018large}, StarGAN~\cite{choi2018stargan}, and CycleGAN~\cite{zhu2017unpaired}. This collection of real and fake samples has been used by subsequent detection methods to train and evaluate models~\cite{ojha2023towards, corvi2023detection}, but since it lacks newer generation methods, new custom evaluations appear often~\cite{wang2023dire, lorenz2023detecting, lu2024towards}. Standalone efforts have also been made to curate synthetic image datasets~\cite{wangDiffusionDBLargescalePrompt2022, bird2024cifake}, which can be used to train detectors. However, while there have been attempts to standardize deepfake video detection benchmarks (see Section \ref{datasets:video}), there are no commonly established evaluations for manipulated image detection.

\subsection{Deepfake Audio Datasets and Detection Challenges} 

As synthetic audio becomes increasingly accessible, interest in synthetic speech detection has grown~\cite{yi2023audio, almutairi2022review, Hamza2022}. This trend has spurred the creation of competitions like the ASVspoof Challenge \cite{wang2020asvspoof, yamagishi2021asvspoof}, which aims to reduce reliance on specific knowledge of speaker verification systems and spoofing attacks, encouraging a more realistic evaluation scenario reflecting real-world examples, and the Audio Deepfake Detection (ADD) Challenge~\cite{yi2022add, yi2023add}, which aims to identify and analyze deepfake speech utterances, tackling the growing threats from advancements in speech synthesis and VC technologies.

Beyond standarized challenges, research on audio deepfake detection has been energized by the introduction of various datasets. For example, the ``In-the-Wild'' dataset~\cite{müller2022does} contains audio deepfakes featuring politicians and public figures gathered from the internet. The ASVspoof DF dataset~\cite{wang2020asvspoof, yamagishi2021asvspoof} comprises genuine and spoofed speech recordings that have been subjected to various lossy codecs such as m4a that are commonly employed in media storage. The DEEP-VOICE dataset~\cite{bird2023realtime} curates authentic human speech recordings from eight notable individuals, along with voices altered to mimic each other through Retrieval-based VC techniques. Similarly, datasets like the Fake or Real (FoR) dataset~\cite{8906599}, WaveFake~\cite{frank2021wavefake}, EmoFake~\cite{zhao2023emofake},~SceneFake \cite{yi2024scenefake}, and H-Voice~\cite{BALLESTEROS2020105331} offer specialized perspectives, each contributing uniquely to the field. The FoR Dataset comprises over 198,000 utterances from deep-learning speech synthesizers and real speech, serving as a cornerstone for research in speech synthesis and synthetic speech detection. WaveFake provides samples from various network architectures and languages; EmoFake explores the impact of altering audio emotions; and SceneFake aims to detect manipulated audio scenes through modifications of the acoustic environment. Lastly, H-Voice consists of 6,672 histograms derived from authentic and synthetic voice recordings. 

\subsection{Deepfake Video Datasets and Detection Challenges}\label{datasets:video}

One of the earliest and most influential datasets in the domain of deepfake video detection is the FaceForensics++ dataset~\cite{roessler2019faceforensicspp}. It includes over 1,000 real video sequences collected from YouTube and corresponding deepfakes created using four different manipulation methods: Deepfakes~\cite{FaceSwap}, Face2Face~\cite{thies2016face2face}, FaceSwap~\cite{marekkowalskifaceswap_2024}, and NeuralTextures~\cite{thies2019deferred}. The dataset is widely used for training and benchmarking detection algorithms due to its diversity in manipulation techniques, but remains limited by the small number of unique identities present in the videos. Other early works include the WildDeepfake dataset~\cite{zi2020wilddeepfake}, which contains a set of 7314 sequences of deepfake content collected from the internet, and the DeepfakeTIMIT dataset~\cite{korshunov1812deepfakes}, a collection of public real videos each with a GAN generated deepfake counterpart.

Building on the foundational work of datasets like FaceForensics++, the DeepFake Detection Challenge (DFDC) dataset was introduced by Facebook AI to spur advancements in detection technologies~\cite{dolhansky2020deepfake}. The DFDC dataset is one of the most comprehensive public datasets, containing over 100,000 face swap video clips of both real and manipulated content. It is notable for having faces of more than 3,000 subjects and using 8 facial modification algorithms. It uses several deepfake generation techniques including GAN-based and non-learned methods to produce the clips. 

Another significant contribution is the Celeb-DF dataset~\cite{li2020celebdf}, which addresses some of the limitations found in earlier datasets such as visual artifacts. Celeb-DF contains 590 YouTube videos of celebrities of varying gender, age, and ethnicity, and 5,639 high-quality deepfake videos of these subjects generated using improved deepfake algorithms that reduce common visual artifacts, thereby posing a greater challenge for detection systems. 

In addition to these datasets, other efforts have focused on generating more complex and varied datasets. For instance, the DeeperForensics-1.0 dataset~\cite{jiang2020deeperforensics10} introduces perturbations such as compression and noise to its video sequences to create a more rigorous and comprehensive benchmark. It contains over 17 million frames and 11,000 manipulated deepfake videos generated using DeepFake VAE swapping. This dataset is subjected to a wide range of real-world distortions, resulting in a larger and more diverse collection of face swap videos that better reflect the complexities of real-world scenarios.

AV-Deepfake1M~\cite{cai2023avdeepfake1m} is one of the largest deepfake datasets to date, containing over 2,000 subjects and 1 million videos with various types of manipulations, including video, audio, and audio-visual deepfakes. Before this dataset, there had been limited datasets including small segments of audio-visual manipulations embedded within real videos, leaving detectors susceptible to these types of attacks.

\section{Discussion}
\label{discussion}
The development of deepfake technologies has significantly advanced, leading to a continuous tug-of-war between generation techniques and the corresponding detection methods. This dynamic interplay shapes the landscape of both fields.

\subsection{Challenges in Current Detection Approaches}

\textbf{Data Scarcity and Bias.} One major issue is the lack of comprehensive and diverse datasets that reflect the full range of manipulations generation models can produce. This scarcity leads to detection models that may perform well on specific types of deepfakes but fail to generalize to out-of-distribution media, including new or slightly different deepfake approaches.

Detection datasets are often custom-made, focusing on outputs from specific generation models. This approach allows detectors to identify unique artifacts but limits their ability to generalize. Additionally, relying on custom datasets complicates the direct comparison of detection approaches. Recent efforts such as DeepfakeBench \cite{yan2023deepfakebench} have made strides towards standardizing comparisons between models by creating a benchmark incorporating many different datasets and streamlining the evaluation of models through a series of analysis tools. However, the rapid advancements in generators make detection datasets become quickly outdated.

\textbf{Evolving Generation Techniques.} As generation methods evolve, they often develop capabilities to bypass specific detection mechanisms, especially those relying on detecting artifacts or inconsistencies that newer models no longer produce: the frequency space artifacts characteristic of GAN-generated images are noticeably absent from diffusion-generated images~\cite{corvi2023detection} and adversarial perturbations can evade detectors~\cite{carlini2020evading}. Relatedly, the visibly increased resolution and realism of deepfakes make it challenging for human experts and automated systems to distinguish between genuine and manipulated content. This raises concerns about the effectiveness of current detection technologies and how to best curate datasets if the authenticity of media cannot be easily verified.

\subsection{Future Directions}

\textbf{Curate Robust Datasets and Design Competitions for Detection.} To effectively combat deepfakes, there is a critical need for creating and maintaining robust, diverse, and representative datasets that are publicly available to the research community. These datasets should include a wide variety of deepfake types and techniques to better train and test detection models. Diversity in modality should also be present; these datasets should include different types of media such as videos, audio, and images from various demographics and in multiple languages to ensure comprehensive coverage. The datasets should be continuously updated at a regular cadence to include the latest deepfake techniques and real-world examples, ensuring that detection models can learn to counter new threats. By updating datasets in this way, deepfake detection methods can be trained more closely to detect the in-the-wild examples that pose the greatest threats to mislead individuals. Creating competitions for model evaluations can also establish standardized comparisons that test detectors on these pertinent examples as well.

\textbf{Focused Efforts on Representation and Consent.} The use of web-scraped data to train both generation and detection models also raises concerns about representation and consent. Deepfake subjects are often public figures like celebrities since ample data is available for them online. However, as the barrier to create deepfakes lowers, issues around non-consensual deepfake pornography and identity misuse will likely become more prevalent. While such content should be moderated and prevented from being created through policy and moderation considerations, detection methods must also be proactive in handling such cases. 

\textbf{Harness Capabilities of Foundation Models for Deepfake Detection.} One avenue parallel to curating robust datasets for training detection models is to harness the capabilities of foundation models that have been pretrained on large amounts of data, likely already inclusive of deepfake datasets. Some detection methods have started to utilize pretrained CLIP~\cite{radford2021learning} models~\cite{ojha2023towards, pianese2024trainingfree, tariang2024synthetic} to detect both fake images and audio, but these methods are relatively new and still face the same difficulties when it comes to standardized evaluations.

\section{Conclusion}
\label{conclusion}
We provide a short survey of the current landscape of deepfake video generation and detection, as well as of the datasets used to train and evaluate these methods. Although great progress has been made in deepfake video detection, there is still much room for improvement in curating datasets that enable the development of robust, generalizable, and responsible detectors. Closer collaboration between the generation and detection communities could help anticipate future directions and ensure that detectors are well-equipped to handle the latest deepfake techniques. Regular detection competitions using shared and frequently updated datasets are a promising avenue to align research efforts to minimize deepfake misinformation spread and related consequences.



\bibliography{main}
\bibliographystyle{icml2024}



\end{document}